\documentclass[sigconf]{acmart}
\AtBeginDocument{%
  \providecommand\BibTeX{{%
    \normalfont B\kern-0.5em{\scshape i\kern-0.25em b}\kern-0.8em\TeX}}}

\copyrightyear{2021}
\acmYear{2021}
\setcopyright{acmcopyright}\acmConference[CIKM '21]{Proceedings of the 30th ACM
International Conference on Information and Knowledge Management}{November
1--5, 2021}{Virtual Event, QLD, Australia}
\acmBooktitle{Proceedings of the 30th ACM International Conference on Information
and Knowledge Management (CIKM '21), November 1--5, 2021, Virtual Event, QLD,
Australia}
\acmPrice{15.00}
\acmDOI{10.1145/3459637.3482108}
\acmISBN{978-1-4503-8446-9/21/11}

\usepackage{enumerate}
\usepackage{graphicx} 
\usepackage{amsmath}
\usepackage{multirow} % merge cells
 
\usepackage{amsmath, amssymb, mathrsfs} % math
\usepackage{booktabs} % set line space
\usepackage{CJKutf8}
\usepackage{tabu} % beautify table
\usepackage{balance}
\usepackage[symbol]{footmisc}

\acmSubmissionID{2498}

\begin{document}
\fancyhead{}

\title{Fine-Grained Element Identification in Complaint Text of Internet Fraud}

% %%
% %% The "author" command and its associated commands are used to define
% %% the authors and their affiliations.
% %% Of note is the shared affiliation of the first two authors, and the
% %% "authornote" and "authornotemark" commands
% %% used to denote shared contribution to the research.
% \affiliation{%

\renewcommand{\thefootnote}{\fnsymbol{footnote}}

\author{Tong Liu\textsuperscript{\rm 1}\footnotemark[1], Siyuan Wang\textsuperscript{\rm 1}\footnotemark[1]\footnotetext[1]{Tong Liu and Siyuan Wang have equal contribution.}, Jingchao Fu\textsuperscript{\rm 1}, Lei Chen\textsuperscript{\rm 1}, Zhongyu Wei\textsuperscript{\rm 1}\footnotemark[2]\footnotetext[2]{Corresponding author.}, \\Yaqi Liu\textsuperscript{\rm 2}, Heng Ye\textsuperscript{\rm 2}, Liaosa Xu\textsuperscript{\rm 2}, Weiqiang Wang\textsuperscript{\rm 2}, Xuanjing Huang\textsuperscript{\rm 1}}
\affiliation{%
  \institution{\textsuperscript{\rm 1}School of Data Science, Fudan University, China; \textsuperscript{\rm 2}Ant Group, China}
  \country{}
}
\email{{18210980056,wangsy18,chenl18,zywei,xjhuang}@fudan.edu.cn}
\email{fuuuuugcn@gmail.com;{yaqiliu.lyq,daokun.yh,liaosa.xls,weiqiang.wwq}@antgroup.com}

% \affiliation{%

\renewcommand{\shortauthors}{Trovato and Tobin, et al.}
% \renewcommand{\thefootnote}{\fnsymbol{footnote}}
% \footnotetext[2]{Equal contribution.}

\begin{abstract}
Existing system dealing with online complaint provides a final decision without explanations. We propose to analyse the complaint text of internet fraud in a fine-grained manner. 
% to support the decision making of officers and
Considering the complaint text includes multiple clauses with various functions, we propose to identify the role of each clause and classify them into different types of fraud element. We construct a large labeled dataset originated from a real finance service platform. We build an element identification model on top of BERT and propose additional two modules to utilize the context of complaint text for better element label classification, namely, global context encoder and label refiner. Experimental results show the effectiveness of our model. \end{abstract}


\begin{CCSXML}
<ccs2012>
   <concept>
       <concept_id>10010405.10003550.10003557</concept_id>
       <concept_desc>Applied computing~Secure online transactions</concept_desc>
       <concept_significance>500</concept_significance>
       </concept>
 </ccs2012>
\end{CCSXML}

\ccsdesc[500]{Applied computing~Secure online transactions}

\keywords{complaint text mining, fraud element identification, dataset}


\maketitle

\section{Introduction}
\label{intro}

% The rapid growth of Internet economy has greatly enriched people’s life and the consumer-oriented business provides people with much more customized service and effective communication. 
% Due to the easy and handy connection, people are more inclined to share and report their complaints online. 
% Although most complaints are harsh and negative, they are valuable for companies and governments to understand incident mechanism, capture individual demands and avoid potential risks \cite{brennan2017consumer,ferri2018evolving}.
% In the field of e-commercial business, 
% despite the convenience of transfer brought by mobile applications, 
% the user information holds a risk of leakage and tends to be taken advantage by scam artists. 

In the field of e-commercial business, various payment platforms provide an easy way of capital transferring but also exposes huge threat of Internet fraud. 
% Every year, the Risk Control Center of financial companies and relevant government departments will receive thousands of fraud complaints, from imprudence remittance to mendacious business contract. 
Every year, financial companies receive thousands of fraud complaints, from imprudence remittance to mendacious business contract. Although complaint text is usually full of sore and loss, it is valuable for organizations to understand incident mechanism and avoid potential risks \cite{brennan2017consumer,ferri2018evolving}. 
% Handling fraud complaints at once not only resolves victims' burning issue with bringing the outlaws to justice, but also helps financial organizations and other law enforcement agencies detect security flaws and thus put an end to unfair and misleading swindle practices.
Handling fraud complaints is of necessity to resolve victims' burning issues and recognizing fraud elements enables financial organizations and law enforcement agencies to detect security flaws and put an end to unfair and misleading swindle practices.   

However, processing numerous complaint texts throws huge challenge for Internet finance companies. 
Different from credit and trading record, fraud complaints are composed of informal and unstructured text that takes painstaking efforts to understand and analyze. 
% On the other hand, in many cases, given a paragraph of complaint text, inspectors usually quickly look through the paragraph and subjectively return a short inspect outcome (accept further investigation or not), which makes many victims indignant for they cannot figure out why their loss never comes back.
Besides, the inspect outcome is usually short (accept further investigation or not), which might make victims indignant for they cannot figure out why their loss never comes back. 
\vspace{-1mm}
\begin{figure}[!th]
\centering
\includegraphics[width=0.85\columnwidth]{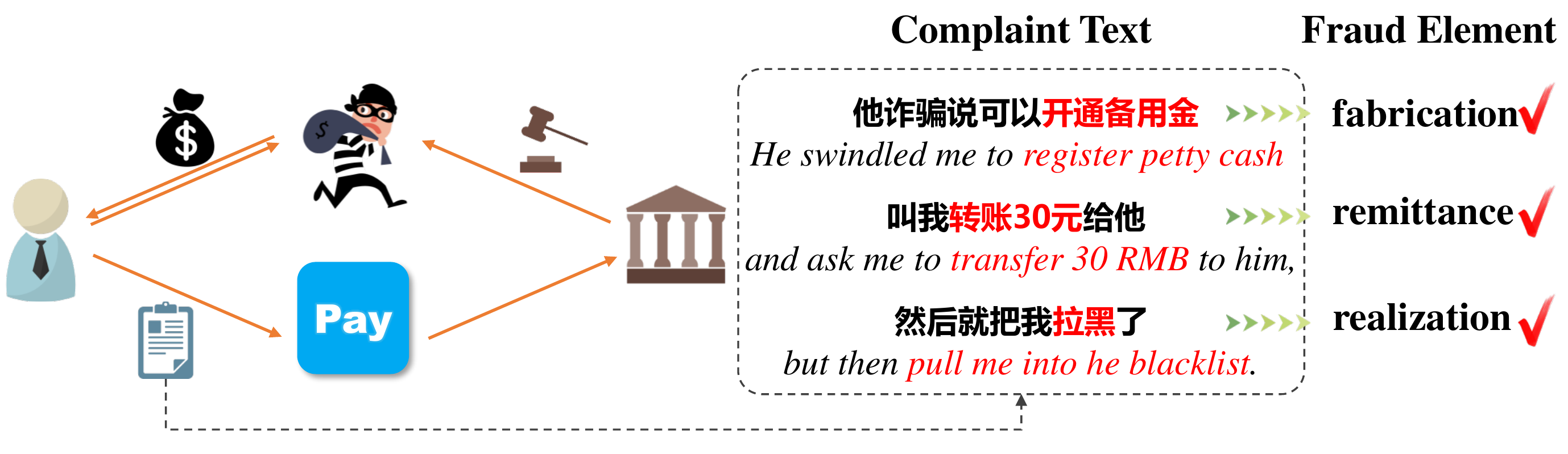}
\vspace{-2mm}
\caption{\label{figure_process} Ideal pipelines of fraud complaints processing.}
\end{figure}

Figure \ref{figure_process} introduces a more well-grounded scheme for complaint text processing and fraud judgement. 
When a user realizes being swindled and submits complaints to the transfer platform, the third-party payment service provider gets involved in and segments the complaint text into pieces so as to inspect fraud elements inside. 
The text on the right briefly shows the three fundamental elements to ascertain a fraud: (1) the fraudster deliberately \underline{\textit{fabricate}} an illusory circumstance to ask for money, (2) the \underline{\textit{remittance}} is material and (3) the injured party \underline{\textit{realizes}} the falsehood afterwards and asks for compensation. 
With fraud elements automatically detected, it is more convenient for inspectors to check whether the statement is valid and more interpretable for users to comprehend the inspect results.
Once the fraud case is established, the third-party and legal institution will follow the hints of fraud elements to investigate the transaction history and then present a fairer arbitrament.

To this end, we propose a novel task to identify fraud elements in a complaint paragraph.
% Based on a fraud complaint corpus provided by an Internet finance service company, we design the annotation criterion and ask undergraduates to annotate each clause. 
We design an annotation criterion, split paragraphs into clauses and ask undergraduates to annotate each clause.
Moreover, we analyze clause distribution in complaint paragraphs and explore the connection between successional clauses while finding that position and global coherence are influential for identifying clause role. 
Therefore, we propose a hierarchical architecture which integrate context information to obtain more accurate prediction of clauses. Our contributions are of three-folds: 

\begin{itemize}
\setlength{\itemsep}{0pt}
\setlength{\parsep}{0pt}
\setlength{\parskip}{0pt}
    % \item We split the complaint text into clauses and propose a novel annotation scheme for fine-grained element identification on clause level to further analyse internet fraud issues.
    \item We propose a novel task and construct a dataset for fine-grained fraud element identification on clause level to further analyse internet fraud issues.
    % \item We build our model on top of BERT and explore to model context information of complaint text to better classify elements. In practice, we design a global context encoder to capture the textual context and a label-refining mechanism to utilize the label context. 
    \item We formulate the task as a form of sequence labeling as plentiful analysis indicates that position and relation are significant features for complaint clauses. 
    % \item We construct a large annotated dataset from real online finance service platform, including 41,103 paragraphs and 197,878 labeled clauses. Experiment results on the dataset show the effectiveness of our proposed model. 
    \item We build our model on top of BERT with a global context encoder to capture the textual context and a label-refining mechanism to utilize the label context. Experiment results on the dataset show the effectiveness of our proposed model. 
\end{itemize}

\section{Dataset}
\subsection{Annotation Framework}
% With the support of an Internet finance service company, we get access to a considerable Chinese complaint corpus collected from its customers. We collaborate to design the annotation criterion which aims at identifying legal and business-related fraud elements.
% Although the practical implementation is much more refined and concrete, t
% The fraud elements can be roughly concluded as 7 categories, including content fabrication(CF), identity fabrication(IF), remittance excuse(RE), contact platform(CP), fraud realization(FR), user demand(UD) and non-fraudulent statement(NONE). 
With the support of an Internet finance service company, we collect a considerable Chinese complaint corpus with 7 categories of fraud elements: \textit{content fabrication} (CF), \textit{identity fabrication} (IF), \textit{remittance excuse} (RE), \textit{contact platform} (CP), \textit{fraud realization}(FR), \textit{user demand} (UD) and \textit{non-fraudulent statement} (NONE). 

The annotation process consists of three steps. First, we split each complaint paragraph into clauses according to Chinese punctuation marks (comma, semicolon and space). 
Second, each clause is assigned to 2 undergraduates to acquire a unique label of fraud element. If any two annotators disagree with each other, the instance will be checked by a third annotator following the majority rule. 
If any two cannot reach an agreement, the instance will be discarded. 
Finally, we calculate the Cohen’s kappa coefficient between two annotators to assess annotation quality and get an averaged kappa coefficient of 78\%. 
Overall, we construct a fraud complaint dataset containing 41,103 paragraphs and 197,878 labeled clauses.

\subsection{Dataset Analysis}
Different from existing complaint-related work \cite{tong2018complaint,hacohen2019automatic,filgueiras2019complaint} that regards every single complaint text as independent, our corpus contains a hierarchical relationship between clauses and paragraphs. Therefore, except for categorical statistics, we further explore the distribution and relation of clauses in specific paragraph.

\paragraph{Categorical Statistics} Table \ref{table_statistics} shows the statistics of each category.
% including quantity proportion and text features. 
% From the perspective of class volume, t
The proportion of each category is quite uneven which makes the task challenging.
% but more representative of the real world. 
\textit{Non-fraudulent statements} account for a large proportion since clauses exclusive of a specific fraud element will be regard as non-fraudulent. 
% Those seemingly irrelevant contents usually contain additional description of the fraud or strong emotion of victims which also make the whole complaint paragraph coherent. 
% Among the 6 other specific categories, 
The elements of \textit{content fabrication}, \textit{remittance excuse} and \textit{fraud realization} have large amounts which are the critical elements for a fraud case. 
The number of \textit{identity fabrication} elements is relatively small, but as an important supplement of factuality modification, they are essential for companies and legal institutions to locate credulous people and scam artists.
Although the elements of \textit{contact platform} and \textit{user demand} only take up a minor space, they are indicative for the third-party companies to determine whether they have the authority of intervention and stimulate them to provide better customer service. 
\vspace{-2mm}
\begin{table}[htb]
% 	\centering
    \setlength{\itemsep}{0.8pt}
	\caption{Statistics of the fraud complaint dataset.}
	\vspace{-1mm}
	\resizebox{0.48\textwidth}{!}{
	\begin{tabular}{c|ccccccc}
	\hline
	\bf Statistics & \bf CF & \bf IF & \bf RE & \bf CP & \bf FR & \bf UD & \bf NONE \\
	\hline
	\# of clauses & 39,207 & 2,739 & 19,546 & 7,882 & 35,289 & 2,608 & 90,607 \\
	Proportion & 19.81\% & 1.38\% & 9.88\% & 3.98\% & 17.83\% & 1.32\% & 45.79\% \\ \hline
	Avg. length of clauses & 12.0 & 12.0 & 11.4 & 10.5 & 9.9 & 10.0 & 8.5 \\
	Vocabulary size & 12,646 & 2,337 & 6,041 & 3,597 & 7,382 & 1,595 & 17,067 \\
	Clause novelty & 0.323 & 0.853 & 0.309 & 0.456 & 0.209 & 0.612 & 0.188 \\
	\hline
 	\end{tabular}
 	}
	\label{table_statistics}
\end{table}

% As for the text attributes of clauses, the analysis mainly focuses on semantic richness.
% Besides, we count how many Chinese characters appear in a clause and calculate the average length of clauses in each category to measure their semantic richness. 
Besides, we measure the semantic richness of each clause by calculating their average length.
As shown in Table~\ref{table_statistics}, the elements of \textit{fabricating content} and \textit{identity} are longer which confirms that they contain more abundant information to depict the fraud action.
\textit{Non-fraudulent statements} are the shortest because they are more casual and untargeted. 
% we split each clause into words with the tool named jieba
%\footnote{\url{https://github.com/fxsjy/jieba}}
% we count how many unique words appear to
We also compute vocabulary size for each category and divide the size by the number of clauses to obtain clause novelty.
% To alleviate the impact of class imbalance, we . 
% In this way, the impact of class imbalance is alleviated.
% and clause novelty basically means how many new words appear in clause for the first time. 
Since \textit{identity fabrication} and \textit{user demand} are barely mentioned, their repetitive use of words is quite small. For frequently emerged elements such as \textit{content fabrication} and \textit{fraud realization}, their novelty difference reflects that fraudsters practice various deception, but most of the cheated are suffering alike.
\vspace{-1.5mm}
\begin{figure}[!ht]
\centering
\includegraphics[width=0.9\columnwidth]{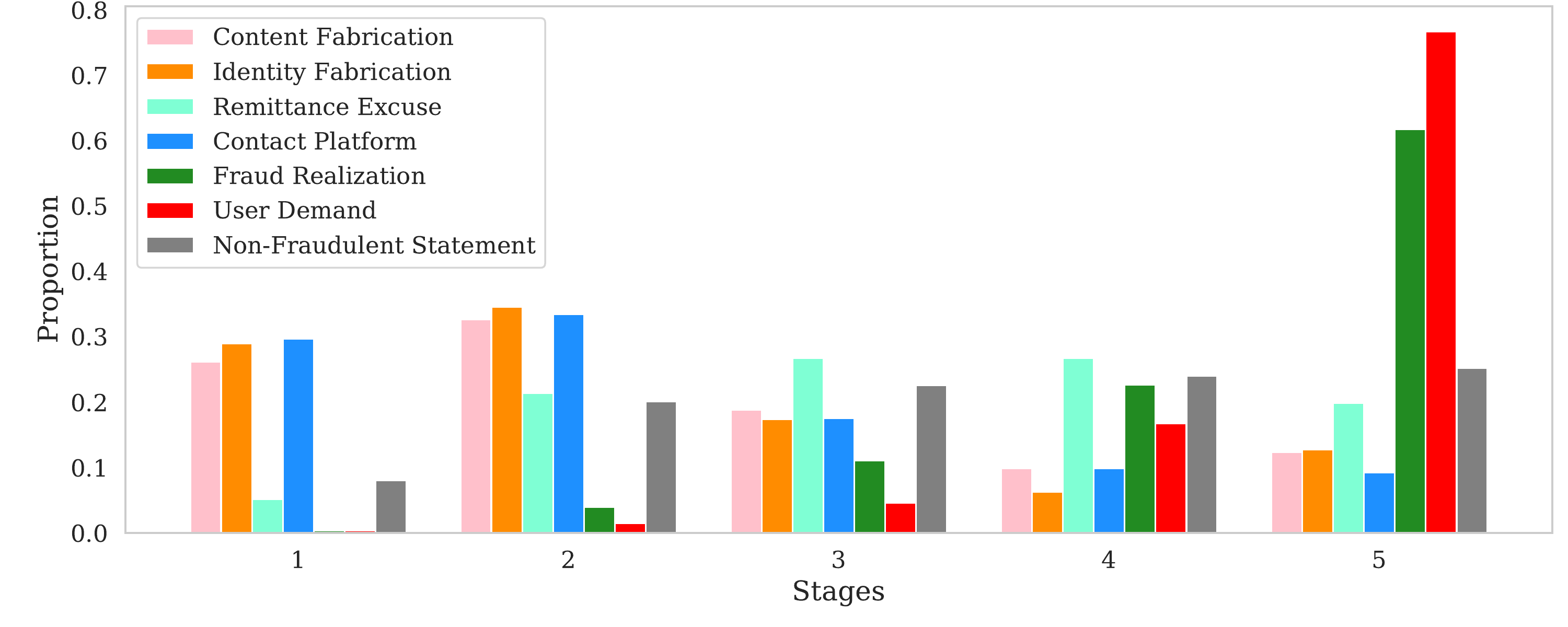}
\vspace{-1mm}
\caption{\label{figure_dist} Distribution of fraud elements in different stages of complaint paragraph.}
\end{figure}
\vspace{-2mm}

\paragraph{Positional Distribution}
Since clauses in the same paragraph are mutually complementary,
% to make the whole submission evidence-sufficient, 
it is worthwhile to investigate
% look into what makes up a qualified fraud complaint and 
how fraud elements distribute among complaint paragraphs.
Given a paragraph, we record the serial number of each clause
% according to the sequential order 
and divide by the sequence length to obtain the relative position. 
Further we segment the whole paragraph into 5 equidistant stages. 
% and assign clauses into the intervals they belong to.
The proportion of different stages for each fraud element is shown in Figure \ref{figure_dist}.
The elements of \textit{content fabrication}, \textit{identity fabrication} and \textit{contact platform} are more likely to appear in the early stage as they come straight to describe the origin, development and transition of the fraud case. The towering bars of \textit{fraud realization} and \textit{user demand} in the rear stages illustrate victims tend to state outcomes and make requests at the end. 
\textit{Non-fraudulent statements} are quite evenly dispersed in different stages for they have the effect of lubrication to make the whole paragraph clear and coherent. 
\vspace{-1.5mm}
\begin{figure}[!th]
\centering
\includegraphics[width=0.9\columnwidth]{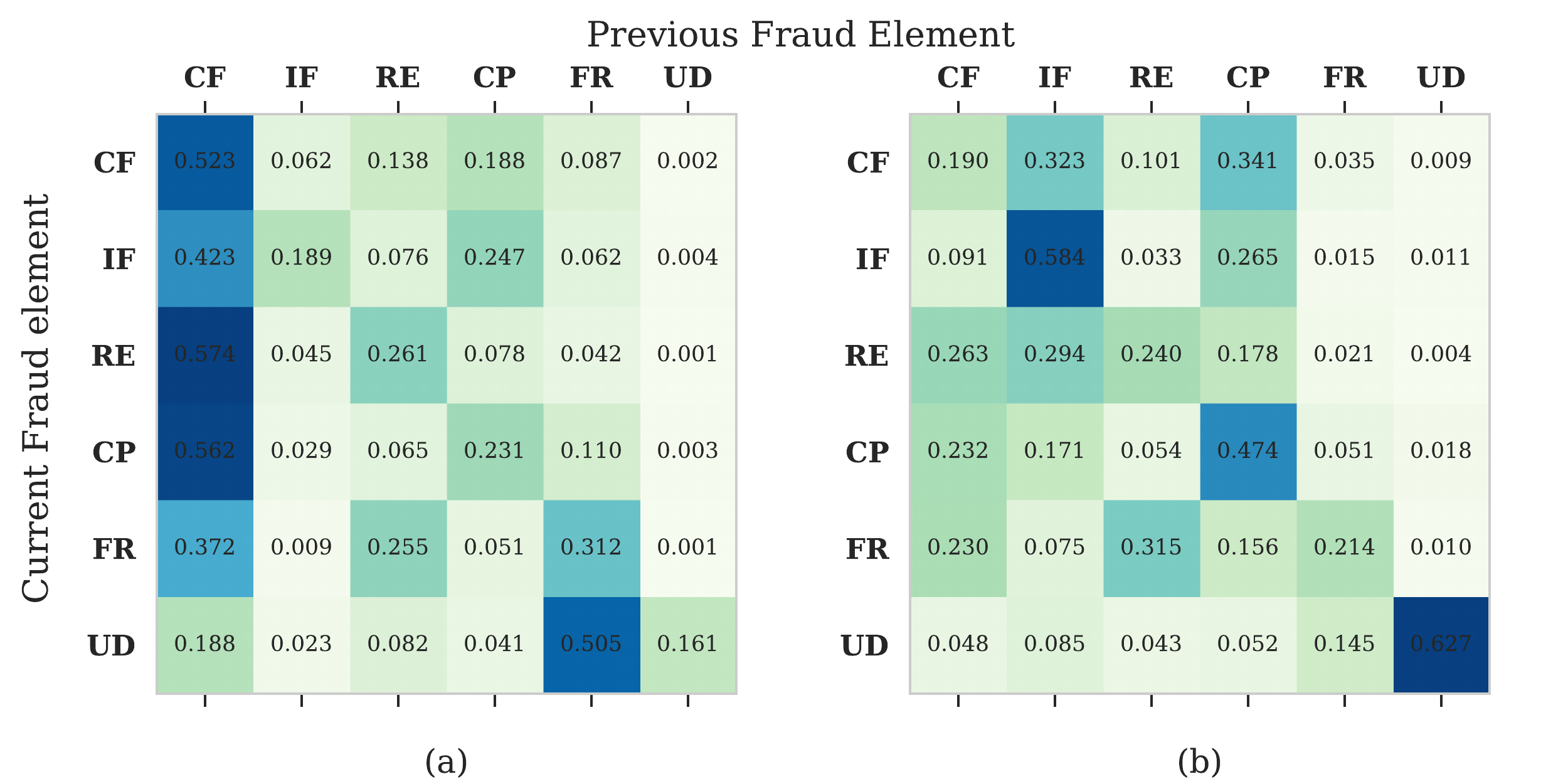}
\vspace{-1mm}
\caption{\label{figure_relation} Probabilities of successional fraud element pairs: (a) the original result; (b) the balanced result.
% considering prior distribution of previous category.
}
\end{figure}
\vspace{-2.5mm}

\paragraph{Ordinal Relation}
% Inspired by N-gram language model \cite{mladenic1998word}, 
We investigate all the possible combinations of successional element pair \cite{mladenic1998word}.
Since non-fraudulent statements lack of explicit semantics, we filter them out. 
For each of other categories, we calculate the proportion of the adjacent parent. 
Similarly, we remove the effect of class imbalance by dividing the prior possibility of the previous fraud element before computing the proportion.
The original and balanced results are shown in Figure \ref{figure_relation}.
Before relieving the imbalance effect, it is inevitable that \textit{content fabrication} is more likely to appear before other elements due to its vast existence. 
Even so, the \textit{user demand} is less likely to follow \textit{content fabrication} which means an absence of consistency between them. 
However, with a nearly same proportion, the \textit{fraud realization} is less possible to become the prerequisite except for {user demand}. 
After normalizing with prior distribution, the elements of \textit{content fabrication}, \textit{identity fabrication} and \textit{contact platform} are the most possible previous statements. Interestingly, both (a) and (b) in Figure \ref{figure_relation} have a deep-colored diagonal indicating that clauses frequently succeed the previous of the same type and there exists a strong semantic transitivity between consecutive clauses.

With all the findings, we claim that the position and relation are significant attributes for complaint clauses, and it is reasonable to treat fraud element identification as a sequence labeling task.

\section{Task and Model}
\begin{figure}[!th]
\centering
\includegraphics[width=0.9\columnwidth]{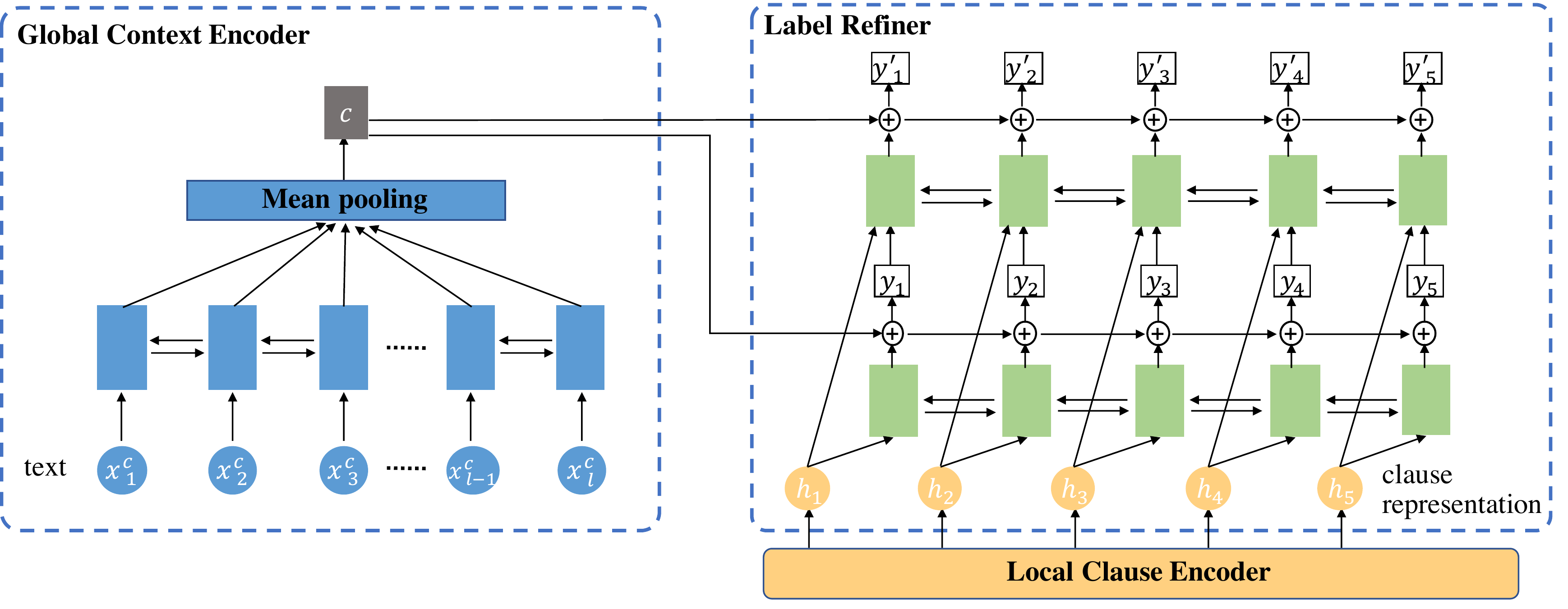}
\vspace{-1mm}
\caption{\label{figure_framework} The overall architecture of our proposed model.}
\end{figure}

This section presents the task of fraud element identification in complaint text and our classification model. The overall framework is shown in Figure~\ref{figure_framework}, which is based on a pre-trained BERT model~\cite{devlin2018bert}.
% which will be introduced in \S~\ref{section_base_model} and 
It consists of a Local Clause Encoder, a Global Context Encoder and a Label Refiner.
% (\S~\ref{section_our_model}). 
The Global Context Encoder is designed to capture the textual context across the complaint text and Label Refiner is proposed to utilize label context.
\vspace{-1mm}
\vspace{-2mm}

\subsection{Task Definition}
We first introduce some notations in our task:
\begin{enumerate}[-]
\itemsep-0.2em
    \item $x=(x_{1},..., x_{n})$: an complaint text with $n$ clauses, where $x_{i}$ is the $i$-th clause;
    \item $x_i=(x_{i,1},..., x_{i,m})$: a clause with $m$ tokens, where $x_{i,j}$ is the $j$-th token in $x_i$;
    \item $y_i$: the fraud element of the $i$-th clause $x_i$ in the complaint text $x$.
\end{enumerate}
We formulate the task as a sequence-level classification task that assigns a element type $y_i$ for each clause $x_i$ in the complaint text $x$. 
% and a sequence labeling task to identify fraud elements across the complaint text $x$.

% \subsection{Base Model: Fine-tuning BERT}
% \label{section_base_model}
% To solve the sequence-level classification task, we base our model on the fine-tuning Bidirectional Encoder Representations from Transformers (BERT)~\cite{devlin2018bert}, which is a task-specific model incorporating BERT with one additional output layer. It take each clause $x_i$ in $x$ as the input of the BERT. Then it feeds the representation of special classification symbol ([CLS]) to the output layer to compute the softmax probability of $x_i$ belonging to different fraud elements as following,
% \begin{align}
%     P(y_i|x_i) = Softmax(W_0 h_{[CLS]})
%     \label{algin_softmax}
% \end{align}
% where $W_0$ is the trainable weight matrix and $h_{[CLS]}$ is the learned representation of [CLS] symbol.

\subsection{Our Model}
\label{section_our_model}
% The architecture of our model is shown in Figure~\ref{figure_framework} which consists 
% Our model is composed of a Local Encoder, a Global Encoder and a Classification Module. 
In our model, the Local Clause Encoder first utilizes BERT to encode each clause in a complaint text $x$ and take the representations of [CLS] symbol as the clause representations $(h_1, ..., h_n)$. Then we introduce the Global Context Encoder and the Label Refiner.

\paragraph{Global Context Encoder} Considering a complaint text contains several fine-grained clauses,
% usually has a specific topic, such as game and loan. 
we introduce a global context encoder~\cite{niu2019enhancing} to capture the global textual context information $c$ to help identify the element of clauses in the complaint text $x$. We concatenate all clauses in the complaint text $x^c=concat([x_{1},..., x_{n}])$ and take the embeddings of them as the input of the global encoder, and use a bi-directional GRU~\cite{huang2015bidirectional} to encode the whole text sequence. Then a mean pooling is conducted over encoding states of all time steps to get the textual context representation as Eq.(\ref{expression_context}),
where $E_w$ is the word embedding matrix.
\begin{equation}
    \begin{normalsize}
    % \begin{small}
        \begin{aligned}
        h^c_i &= {\rm BiGRU}(E_w x^c_i, h^c_{i-1}) \\ 
        c &= {\rm mean}(h^c_1, ..., h^c_i,...)
        \label{expression_context}
        \end{aligned}
    % \end{small}
    \end{normalsize}
\end{equation}

\paragraph{Label Refiner}
After encoding the clauses and getting the textual context representation of the complaint text, we identify the fraud element of each clause. Instead of directly feeding [CLS] representation from BERT into a output layer, 
% being aware of context information of each clause in the complaint text is necessary for clause classification, so 
we take the clause representations $(h_1, ..., h_n)$ in the complaint text $x$ as the input and employ a BiGRU to learn the dependencies between the clauses and get their hidden states as $(s^1_1, s^1_2,..., s^1_n)$. To be aware of the global textual context information, we concatenate the state $s^1_i$ of clause $x_i$ with the textual context representation $c$~\cite{tanaka2019dialogue} to get the classification probability through a feed-forward network as Eq.(\ref{expression_softmax_2}).
% where $W_1$ is the trainable weight matrix.
\begin{equation}
    \begin{normalsize}
    % \begin{small}
        \begin{aligned}
        %  &= {\rm BiGRU}(h^i, s^1_{i-1}) \label{expression_bigru} \\
        P(y_i|x_i, x) &= Softmax(W_1 [s^1_i;c])
        \label{expression_softmax_2}
        \end{aligned}
    % \end{small}
    \end{normalsize}
\end{equation}

Then we propose a \emph{Label-Refining} mechanism to utilize the label context information
% We assume that after the first classification, again classifying each clause into a fraud element based on the first classification result can help
to refine the identification results. As shown in Figure~\ref{figure_framework}, after the first BiGRU layer for classification, we take the concatenation of each clause representation $h_i$ and the previous classification probability $P(y_i|x_i, x)$ 
% of the clause $x_i$ 
as the input and employ another BiGRU to again encode the clauses sequence in the complaint text $x$. Then we get the refined probability $P(y'_i|x_i, x)$ as Eq.(\ref{expression_softmax_3}).
% and $W_2$ is a trainable weight matrix. 
% \begin{align}
%     s^2_i &= {\rm BiGRU}([P(y_i|x_i, x); h^i], s^2_{i-1}) \label{expression_bigru_2} \\
%     P&(y^{'}_i|x_i, x) = Softmax(W_2 [s^2_i;c])
%     \label{expression_softmax_3}
% \end{align}
\begin{equation}
    \begin{normalsize}
    % \begin{small}
        \begin{aligned}
        s^2_i &= {\rm BiGRU}([P(y_i|x_i, x); h^i], s^2_{i-1}) \\
        P&(y'_i|x_i, x) = Softmax(W_2 [s^2_i;c])
        \label{expression_softmax_3}
        \end{aligned}
    % \end{small}
    \end{normalsize}
\end{equation}

\section{Experiments}
\subsection{Experimental Setup}
We randomly split the dataset into training, validation and test set by the proportion of 8:1:1. The numbers of complaint paragraphs and clauses in each data split are (32882, 4110, 4111) and (158521, 19731, 19626), respectively. 

We adopt BERT-base, Chinese model as backbone. 
% and Tencent AI Lab Embedding Corpus for Chinese Words and Phrases\footnote{https://ai.tencent.com/ailab/nlp/zh/embedding.html} is used to initialize the word embedding in the global context encoder with dimension 200. 
The global encoder is a two-layer GRU while the GRUs in label refiner are both of one-layer, and the size of hidden units in all GRUs is set as 256.
Mini-batch of size 32 is taken and the dropout rate is 0.3. We first fine-tune BERT for the classification task as the base model for 4 epochs using AdamW optimizer with learning rate of 2e-5. Then we load the fine-tuned BERT as the local clause encoder and freeze its parameters, and then train our model for another 10 epochs using Adam optimizer with learning rate 2e-4.

\subsection{Overall Performance}
We compare \emph{Our Model} with some baseline and state-of-the-art models, including \emph{SVM}~\cite{suykens1999least}, \emph{BiGRU}~\cite{liu2019bidirectional}, \emph{BERT}~\cite{devlin2018bert}, \emph{BERT-wwm-ext}~\cite{cui2019pre}, \emph{BERT+BiGRU} and \emph{RoBERTa}~\cite{liu2019roberta}. \emph{Our Model-GC} and \emph{Our Model-LR} are ablation tests of our model, without the global context encoder and the \emph{label-refining} mechanism,  respectively. 
\vspace{-1mm}
\begin{table}[ht]
% \renewcommand\arraystretch{1.1}
% 	\centering
	\caption{Evaluation results for element identification in complaint text of different models.}
	\vspace{-1mm}
	\label{table_result}
	\resizebox{0.48\textwidth}{!}{
	\begin{tabular}{c|cccc}
	\hline
	\bf Model & \bf Accuracy(\%) & \bf Precision(\%) & \bf  Recall(\%) & \bf F1-score(\%) \\
	\hline
	\emph{SVM} & 77.22 & 79.38 & 71.41 & 75.19 \\
	\emph{BiGRU} & 80.51 & 79.46 & 77.31 & 78.37  \\
	\emph{BERT} & 83.53 & 83.66 & 80.64 & 82.13 \\
	\emph{BERT-wwm-ext} & 83.27 & 82.80 & 81.63 & 82.21 \\
	\emph{BERT+BiGRU} & 82.70 & 82.38 & 80.44 & 81.40 \\   
	\emph{RoBERTa} & 83.40 & 82.39 & 81.97 & 82.18 \\
% 	BERT+CRF & 83.50 & 82.75 & 82.27 & 82.51 \\
	\hline
	\emph{Our Model-LR} & 84.11 & 83.28 & 82.31 & 82.79 \\
	\emph{Our Model-GC} & 84.30 & 83.73 & \bf 82.35 & 83.04 \\
	\emph{Our Model} & \bf 84.47 & \bf 84.11 & 82.30 & \bf 83.19  \\
	\hline
 	\end{tabular}
 	}
\end{table}

We take Accuracy, Precision, Recall, and F1-score as evaluation metrics, and results are shown in the Table~\ref{table_result}.
% for automatic evaluation of classification. 
We have several findings as follows: (1) \emph{Our Model} outperforms others in terms of all metrics except recall. This indicates the effectiveness of our model considering both textual context and label context information. (2) \emph{BiGRU} performs worse than all other BERT-based models, which demonstrate the effectiveness of pre-trained models. (3) \emph{BERT}, \emph{BERT-wwm-ext} and \emph{RoBERTa} achieve similar performance, which means that \emph{BERT} is enough to capture clause representation for classification. Thus we adopt \emph{BERT} as our base model. (4) The improvements of \emph{Our Model} compared to \emph{Our Model-LR} and \emph{Our Model-GC} respectively reveal the effectiveness of the global context encoder and \emph{label-refining} mechanism for our element identification task.
% \begin{enumerate}[-]
% \setlength{\itemsep}{1pt}
% \setlength{\parskip}{1pt}
% % \itemsep-0.1em
%     \item Our model outperforms other models in terms of all metrics except recall. This indicates the effectiveness of our proposed model considering both textual context and label context information.
%     \item BiGRU performs worse than all other BERT-based models, which demonstrate the effectiveness of pre-trained language model for the sequence-level classification task.
%     % \item BERT+fine-tuning performs better than BERT+BiGRU which freezes the BERT parameters. It shows the necessity of fine-tuning of pre-trained language model for downstream classification. 
%     \item BERT+fine-tuning, BERT-wwm-ext+fine-tuning and RoBERTa+fine-tuning achieve similar performance, which means that BERT model is enough to capture clause representation for classification. Thus we adopt the BERT+fine-tuning as our base model. 
%     \item In ablation tests, the improvements of Our Model compared to Our Model-\emph{LR} and Our Model-\emph{GC} respectively reveal the effectiveness of the global context encoder and \emph{label-refining} mechanism for our element identification task.

% \end{enumerate}

\subsection{Further Analysis}
\paragraph{Results of different fraud element} We investigate the performance of different fraud element types of our model in terms of Precision, Recall and F1-score in Table~\ref{table_result_2}. We can see that for clauses of Fraud Realization, User Demand and Non-Fraudulent Statement, our model achieves a good performance.
From the Figure~\ref{figure_dist} which shows clauses of Fraud Realization and User Demand concentrated in the last stage of the complaint text, our model is easier to capture this pattern so that it performs better for these two elements. As for Non-Fraudulent Statement, we assume the good performance is owing to its large proportion in the dataset.
\vspace{-1mm}
\begin{table}[ht]
	\caption{Evaluation results for complaint text classification of different fraud element types.}
	\label{table_result_2}
	\vspace{-1mm}
	\resizebox{0.45\textwidth}{!}{
	\begin{tabular}{c|ccc}
	\hline
	\bf Fraud Element & \bf Precision(\%) & \bf Recall(\%) & \bf F1-score(\%) \\
	\hline
	CF & 82.68 & 82.33 & 82.50  \\
	IF & 83.90 & 82.49 & 83.19 \\
	RE & 75.85 & 63.74 & 69.27 \\
	CP & 85.90 & 83.23 & 84.54 \\
	FR & 86.07 & \bf 89.16 & 87.59 \\
	UD & \bf 88.43 & 86.99 & \bf 87.70 \\
	NONE & 85.92 & 88.15 & 87.02 \\
	\hline
 	\end{tabular}
 	}
\end{table}
\vspace{-1mm}

\paragraph{Error Analysis} We compute the confusion matrix to analyse which pairs of fraud elements are more confused to be classified. We also compare the confusion matrices of \emph{Our Model-LR} and \emph{Our Model} to see which elements are improved by the label refining mechanism.
\begin{figure}[ht]
\centering
\includegraphics[width=0.9\columnwidth]{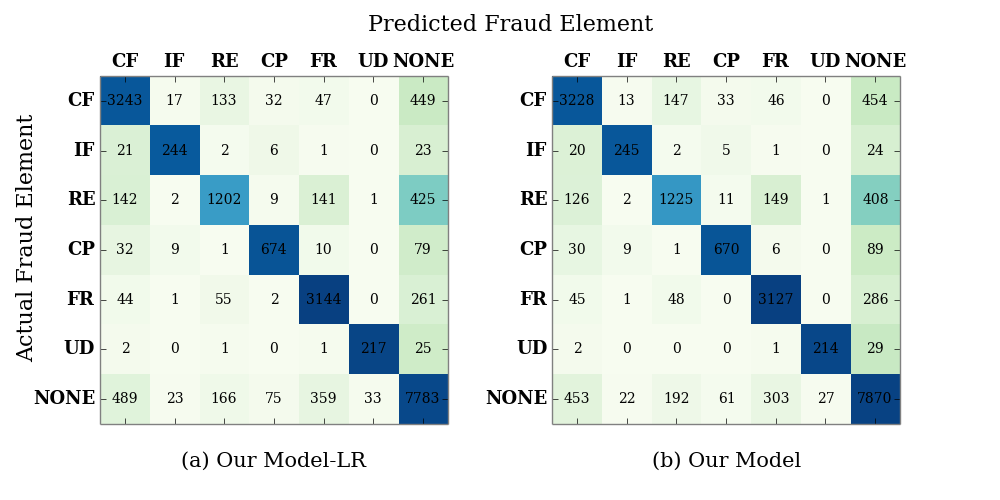}
\caption{\label{figure_confusion} The confusion matrices for classification results of \emph{Our Model-LR} (a) and \emph{Our Model} (b).}
\end{figure}
As shown in Figure~\ref{figure_confusion}, we can see that clauses of Non-Fraudulent Statement are easier to be confused with all other elements, because it makes up the majority of the imbalanced dataset. And the pairs of (Identity Fabrication, Content Fabrication), (Remittance Excuse, Content Fabrication) and (Remittance Excuse, Fraud Realization) are likely to be classified incorrectly.
From comparison between the diagonal elements of matrix (a) and (b), we find that the classification of clauses in Remittance Excuse, Non-Fraudulent Statement and Identity Fabrication are indeed improved, and the corresponding confused numbers are decreased, for example, (Remittance Excuse, Content Fabrication) is reduced from 142 to 126.

\section{Related Work}
% \paragraph{Complaint Text Analysis}
Recently many complaint classification tasks have been widely studied \cite{kos2017classification}.
% designed to handle complaint automatically.
% In the era of technology, an increasing number of researchers set about excavating merit from complaint data and designing framework to handle complaint automatically. Since complaint text usually contains sensitive information, most of the corpus is not open to the public, and complaint classification tasks are mostly designed to reduce human cost for service providers, generally seen in financial, commercial and governmental fields. 
% \citet{kos2017classification} collect complaints regarding trade fraud and form a binary classification task to determine whether the fraud case deserves further investigation. Their work mainly concentrates on dealing with skewed data while the process of problem formulation is roughly mentioned.
% In the customer-oriented area, r
Researchers categorize complaints in order to figure out the complaint reasons \cite{tong2018complaint}, identify downstream companies \cite{hacohen2019automatic}, and explore user demands \cite{joung2019customer}.
\citet{filgueiras2019complaint} analyze the complaints about food safety and economic surveillance to help government structuralize complaint letters and decide the fine-grained department ultimately responsible for the complaint. 
In our work, rather than assigning labels to the whole complaint text, we map each component of the paragraph and list all the fraud elements to provide more interpretable judicial outcomes.

% \paragraph{Text Classification}
% Text classification is a classical task of natural language processing.
Traditional methods for text classification utilize manually crafted features, and employ machine learning algorithms such as Naive Bayes \cite{mccallum1998comparison} and Support Vector Machines \cite{cortes1995support} to obtain category bounds. 
% When the dense real valued vectors of words \cite{mikolov2013distributed} are available, it is more flexible to acquire various representation of sentences. 
% Especially with the popularity of deep learning, 
% With the popularity of deep learning, 
% % RNN \cite{gers1999learning,cho2014learning} and CNN \cite{nowlan1995convolutional} are respectively capable of capture sequential information and local N-gram patterns\cite{kim2014convolutional} of words. 
% end-to-end models utilizing RNN and CNN to capture sequential information and local N-gram patterns\cite{kim2014convolutional} of text have shown promising performance.
% Recently, numerous pre-trained models providing context-dependent text embeddings refresh the performance of text classification 
% with large margin using 
% % complex neural networks and 
% large training corpus, such as
% ELMo~\cite{peters2018deep}, ULMFiT~\cite{howard2018universal}, OpenAI GPT~\cite{radford2018improving},
% and BERT~\cite{devlin2018bert}. And RoBERTa~\cite{liu2019roberta} is proposed to
% % improve the result after 
% overcome some limitations of original BERT. 
With the popularity of deep learning, RNN, CNN~\cite{kim2014convolutional}, and pre-trained language models\cite{howard2018universal, radford2018improving, devlin2018bert, liu2019roberta} refresh the performance of text classification.
In this paper, we build our model on top of BERT and propose a fine-grained classification schema for element identification using textual and label context information.

\section{Conclusion}
In this paper, we aim at analyzing complaint text of internet fraud in a fine-grained level. We first propose an annotation scheme to distinguish various types of fraud elements and construct a dataset as benchmark. We build a classification model on top of BERT and use context information of complaint text to better identify element types. Experiment results confirm the effectiveness of our model. 
In the future, we will explore how to utilize domain knowledge for complaint text modeling. We will also utilize the fine-grained element type information for better complaint text classification.

\bibliographystyle{ACM-Reference-Format}
\balance
\bibliography{cikm}

\end{document}